\documentclass[letterpaper, 10 pt, conference]{ieeeconf}  % Comment this line out if you need a4paper

\IEEEoverridecommandlockouts                              % This command is only needed if 
\usepackage{times}  %Required
\usepackage{helvet}  %Required
\usepackage{courier}  %Required
\usepackage{url}  %Required
\usepackage{graphicx}  %Required
\usepackage[export]{adjustbox}
\usepackage{tikz}
\usetikzlibrary{calc}
\usepackage{varwidth}
\usetikzlibrary{shapes,arrows}

\tikzstyle{decision} = [diamond, draw, fill=blue!20, 
    text width=4.5em, text badly centered, node distance=3cm, inner sep=0pt]
\tikzstyle{block} = [rectangle, draw, fill=blue!20, 
    text width=5em, text centered, rounded corners, minimum height=4em]
\tikzstyle{line} = [draw, -latex']
\tikzstyle{cloud} = [draw, ellipse,fill=red!20, node distance=3cm,
    minimum height=2em]

\usepackage{epsfig} % for postscript graphics files
\usepackage{amsmath} % assumes amsmath package installed
\usepackage{amssymb}  % assumes amsmath package installed
\usepackage{amsthm}
\usepackage{thmtools,thm-restate}
\usepackage{newtxtext}%, newtxmath}
\usepackage{xspace}
\usepackage{bbm} % for indicator function
\usepackage{color}
\usepackage{amsfonts}
\usepackage{mathtools}
\usepackage{setspace}
\usepackage{pdfsync}
\usepackage[ruled,vlined, linesnumbered]{algorithm2e}
\SetArgSty{textup}
\usepackage[noend]{algpseudocode}
\usepackage{url}
\usepackage{caption}
\usepackage{subcaption}
\usepackage{bm}
\usepackage{multirow}
\usepackage{booktabs}
\usepackage{xspace}

\graphicspath{{./figs/}}
\newboolean{CONF}
\newboolean{ARXIV}

\setboolean{CONF}{false}
\ifthenelse{\boolean{CONF}}
	{\setboolean{ARXIV}{false}}
	{\setboolean{ARXIV}{true}}

\newcommand{\textVersion}[2]
{\ifthenelse{\boolean{CONF} }{#1}{}\ifthenelse{\boolean{ARXIV}}{#2}{}}

\newboolean{ShowAuthors}
\setboolean{ShowAuthors}{false}
\def\blind#1{
\ifthenelse{\boolean {ShowAuthors}}{#1}{}
}

\newcommand{\stateSpace}[0]{\mathcal{X}}
\newcommand{\Dim}[0]{d}
\newcommand{\state}[0]{x}
\newcommand{\stateSet}[0]{X}
\newcommand{\obsStateSpace}[0]{\stateSpace_{\mathrm{obs}}}
\newcommand{\freeStateSpace}[0]{\stateSpace_{\mathrm{free}}}

\newcommand{\Path}[0]{\xi}
\newcommand{\ind}[0]{\tau}

\newcommand{\graph}[0]{G}

\newcommand{\alg}[0]{\textsc{Alg}}
\newcommand{\denseGraph}[0]{\graph_{\mathrm{dense}}}

\newcommand{\pp}[0]{\Lambda}

\newcommand{\start}[0]{\state_{s}}
\newcommand{\goal}[0]{\state_{g}}

\newcommand{\feat}[0]{y}
\newcommand{\featDim}[0]{m}

\newcommand{\loss}[0]{\mathcal{L}}
\newcommand{\risk}[0]{\mathcal{R}}
\newcommand{\encoderParam}[0]{\phi}
\newcommand{\decoderParam}[0]{\theta}
\newcommand{\encoder}[0]{q_\encoderParam}
\newcommand{\decoder}[0]{p_\decoderParam}

\newcommand{\dataSet}[0]{\mathcal{D}}
\newcommand{\condVar}[0]{\feat}
\newcommand{\latentVar}[0]{z}
\newcommand{\latentDim}[0]{L}
\newcommand{\stateSetVar}[0]{\stateSet}
\newcommand{\stateVar}[0]{\state}

\newcommand{\sparseGraph}[0]{\graph_{\mathrm{sparse}}}

\newcommand{\abs}[1]{\left|#1 \right|}

\newcommand{\bbm}{\begin{bmatrix}}
\newcommand{\ebm}{\end{bmatrix}}

\newcommand{\set}[1]{\left\lbrace #1\right\rbrace}

\newcommand{\Normal}[0]{\mathcal{N}}

\newcommand{\calT}{\ensuremath{\mathcal{T}}\xspace}

\newcommand{\ignore}[1]{}

\newcommand{\gcs}[0]{\textsc{GenerateCriticalSources}\xspace}
\newcommand{\cs}[0]{\textsc{\ensuremath{\mathcal{CS}}\xspace}\xspace}
\newcommand{\ccs}[0]{\textsc{CandidateCS}\xspace}

\newcommand{\legoglobal}[0]{\textsc{LEGO-Global}\xspace}
\newcommand{\legocvae}[0]{\textsc{LEGO$-$CVAE}\xspace}
\newcommand{\expandlocalgraphs}[0]{\textsc{ExpandLocalGraphs}}
\newcommand{\densifylocalgraphs}[0]{\textsc{DensifyLocalGraphs}\xspace}
\newcommand{\randomnode}[0]{\textsc{RandomNode}\xspace}
\newcommand{\nearestvertex}[0]{\textsc{NearestVertex}\xspace}
\newcommand{\connectedcomponent}[0]{\textsc{ConnComp}\xspace}
\newcommand{\cstree}[0]{\textsc{\calT}\xspace}
\newcommand{\localgraph}[0]{\textsc{\ensuremath{\mathcal{LG}}\xspace}\xspace}

\newcommand{\interpolate}[0]{\textsc{Interpolate}\xspace}
\newcommand{\iscollisionfree}[0]{\textsc{isValid}\xspace}
\newcommand{\lcsrrt}[0]{\textsc{LCS$-$RRT}\xspace}
\newcommand{\csrrt}[0]{\textsc{CS$-$RRT}\xspace}
\newcommand{\setln}[0]{\textsc{\ensuremath{\mathcal{LN}}\xspace}\xspace}

\newcommand{\subgraph}[0]{\textsc{Subgraph}\xspace}

\makeatletter 
\let\NAT@parse\undefined
\g@addto@macro{\@algocf@init}{\SetKwInOut{Parameter}{Parameters}} 
\makeatother

\usepackage{hyperref}

\title{\fontsize{15}{12}{\textbf{ 
Robotic Motion Planning using Learned Critical Sources and Local Sampling}}
}

\author{Rajat Kumar Jenamani$^{*1}$, Rahul Kumar$^{*1}$, Parth Mall$^{*1}$ and Kushal Kedia$^{*1}$% <-this % stops a space
\thanks{*All four authors contributed equally to this research.}% <-this % stops a space
\thanks{$^{1}$Indian Institute of Technology Kharagpur \{ rkj, vernwalrahul, parthmall, kushal.k \}@iitkgp.ac.in}%
}

\begin{document}

\setlength{\parindent}{0cm}
\setlength{\textfloatsep}{0pt}
\setlength{\floatsep}{0pt}

\maketitle
\thispagestyle{empty}
\pagestyle{empty}
\begin{abstract}
Sampling based methods are widely used for robotic motion planning. Traditionally, these samples are drawn from probabilistic ( or deterministic ) distributions to cover the state space uniformly. Despite being probabilistically complete, they fail to find a feasible path in a reasonable amount of time in constrained environments where it is essential to go through narrow passages (bottleneck regions). Current state of the art techniques train a learning model (learner) to predict samples selectively on these bottleneck regions. However, these algorithms depend completely on samples generated by this learner to navigate through the bottleneck regions. As the complexity of the planning problem increases, the amount of data and time required to make this learner robust to fine variations in the structure of the workspace becomes computationally intractable. In this work, we present (1) an efficient and robust method to use a learner to locate the bottleneck regions and (2) two algorithms that use local sampling methods to leverage the location of these bottleneck regions for efficient motion planning while maintaining probabilistic completeness.

We test our algorithms on 2 dimensional planning problems and 7 dimensional robotic arm planning, and report significant gains over heuristics as well as learned baselines.
\end{abstract}
\section{Introduction}
We examine the problem of using prior experience for sampling based motion planning in robots. Sampling based motion planning (SBMP) algorithms construct a graph or roadmap as a discrete representation of the state space of a robot. The vertices of this roadmap represent robot configurations and edges represent potential movements of the robot. A graph search algorithm is then used to find the path between any two vertices in the roadmap. Rapidly Exploring Random Tree (RRT)\cite{c17} is a tree growing variant of SBMP that creates this roadmap (tree) implicitly while planning. SBMP is state of the art in high dimensional spaces.

A defining feature of SBMP is it’s reliance on the sampler. Traditionally, these samples are generated either probabilistically or deterministically\cite{c8} to uniformly cover the state space. Such a sampling approach allows arbitrarily accurate representations (in the limit of the number of samples approaching infinity), and thus allows theoretical guarantees on completeness. However, in environments where paths pass through narrow passages, these algorithms become computationally intractable. This is because a huge number of samples must be generated to cover these narrow passages.\begin{figure}[!ht]
    \centering
	\begin{subfigure}[h]{0.7\columnwidth}
		\centering
		  \includegraphics[width=\textwidth]{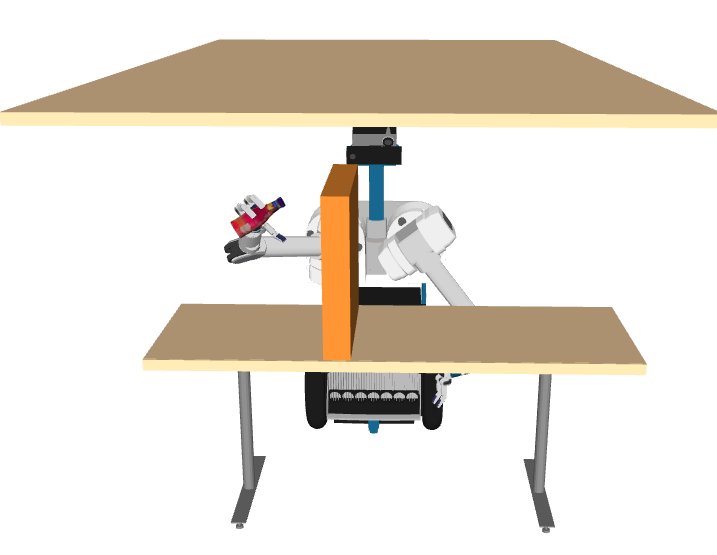}
		  %\caption{}
	\end{subfigure}
	\caption{ A $\mathbb{R}^7$ robotic arm planning problem containing a narrow passage.}
\end{figure}Thus, the main challenge in sampling based algorithms is to place a small set of samples (critical samples) on certain key locations (bottleneck regions) to enable the algorithm to find a high quality path with low computation effort. 
% We call these key locations as bottleneck regions and the samples as critical samples.
% all narrow passages not bottleneck regions

Current state of the art approaches\cite{c1,c2} use learning algorithms (called learners hereon) to predict these critical samples. These samples are then connected to the nodes of a uniform sparse graph to construct a roadmap. 
Depending on the structure of the bottleneck regions (say in an extended narrow passage or semicircular tube like structure), the learner is required to place a small set of critical samples in accordance with the local structure at each bottleneck. 
% In bottleneck regions (say in an extended narrow passage or semicircular tube like structure), 
The graph created by connecting these critical samples among themselves acts as a bridge for connecting the disjoint sub-graphs (connected components) present in the sparse graph. Thus, the learner needs to not only identify the bottleneck regions but also propose samples in accordance with their local structure.

This brings us to two major challenges faced by these approaches. Firstly, the learners used in these approaches are commonly fed an occupancy grid among other parameters as a representation of the workspace of a robot. There exists a trade-off between the size of the occupancy grid and, amount of data and time taken by the model to converge. In complex planning problems, the size of this occupancy grid must be large. However, this makes the convergence of the model computationally intractable. On the other hand, a decrease in the size of the occupancy grid results in low resolution and thus the learner is not able to learn the internal representation of these bottleneck regions. Thus, a large number of the generated samples are rendered useless. Secondly, most of the learners are conditioned only on the planning problem and not on the prior samples it has generated. Thus, the learner tends to repeatedly sample similar points inside a bottleneck region, leading to redundant samples. 

Our key insight is to rely on the learner to identify the location of the bottleneck regions and then exploit the property of planners such as RRT to cover the local structure of these bottleneck regions. Therefore, getting a set of samples from the learner that includes a single sample in each bottleneck region is sufficient. As we expect the learner to generate a single sample per bottleneck region, we propose to convolve the occupancy grid using a kernel to create a smaller grid that contains information relevant for this generation. We use this preprocessed grid during training and for testing the learner. 
% \rvnote{reframe}. 

We thus use the learner to get this initial set of critical samples (one per bottleneck region) and call them critical sources. We then generate our required set of critical samples from these critical sources using local sampling-based methods. 

This solves the first challenge faced by current algorithms, as using a smaller grid as input to the learner makes it converge faster to generalizing well over planning problems that have similar global structure for the location of bottleneck regions but different local structures within these regions 
% \rvnote{long and a little confusing}. 
% \rvnote{people may argue that we don't care about the training time, maybe a better way is to say is ki this enable the model to survive local variations in the structure of the environment}. 
This also solves the second challenge as we condition the subsequent samples on our current set of critical samples. The job is therefore divided between the learner (generate the critical sources) and the local sampler (procure the required set of critical samples by using these critical sources to generate subsequent samples). We argue that this is similar to how we, as humans, given a planning problem, first \emph{globally} identify the doorways present relevant to our planning problem and then \emph{locally} explore how to navigate through these doorways.

Due to the efficient space filling nature of RRTs, we propose the algorithm: Local Critical Source RRT (\lcsrrt). This algorithm, in conjunction with a sparse graph, uses RRTs rooted at critical sources as the local sampling algorithm.

We propose another algorithm, Critical Source-RRT ($\csrrt$), which forfeits the sparse graphs altogether and incrementally builds RRTs rooted at start, goal and critical sources. In addition to RRTs generating the required set of critical samples, $\csrrt$ depends on the inherent bias of RRTs to grow towards large unsearched areas of the problem to emulate the sparse graph once the tree is out of the bottleneck region.

We thus make the following contributions:
\begin{enumerate}
  \item We present $\gcs$, a fast methodology that uses learning models efficiently to generate a set of critical samples, ideally  one per bottleneck region (called critical sources).
  \item We present the algorithm Local Critical Source - RRT ($\lcsrrt$) which, in conjunction with a sparse graph, uses RRTs rooted at critical sources to generate the required set of critical samples which acts as a bridge between disjoint sub-graphs of the sparse graph.
  \item We present the algorithm Critical Source - RRT ($\csrrt$) that works by incrementally building RRTs rooted at multiple key sources (start, goal and critical sources).
  \item We show that $\lcsrrt$ and $\csrrt$ outperform the sampling based baselines on a set of $\mathbb{R}^2$ point object and $\mathbb{R}^7$ robotic arm motion planning problems, and prove their probabilistic completeness.
\end{enumerate} 

% We show that LCS-RRT and CS-RRT outperform the state of the art on different sets of environments and prove the probabilistic completeness of these algorithms. 

% The rest of the paper is arranged as follows, Section I 
% Local Planner instead of sampler?
\section{Related Work}

% Uniform Sampling is bad
An analysis on the shortcomings of using uniform sampling in the presence of narrow passages is given by D. Hsu et. al. \cite{c7} . This has stimulated numerous works on using selective densification for non-uniform sampling \cite{c9}-\cite{c13}.

% Non-Uniform sampling for constructing roadmaps
A multitude of these works use adaptive sampling for roadmap densification by exploiting the structure of the environment. While some propose to sample around the obstacles \cite{c14}, \cite{c15}, others use heuristic based strategies to trace or locate key samples \cite{c9}. Even though these techniques generalize to a large set of problems, they suffer from placing samples in regions where a path is unlikely to traverse. Also, owing to the huge number of collision checks performed, their computation time increases rapidly with increase in dimensionality of the state space. 

% Heuristic Based Planners like Informed-RRT*
% Other works \cite{I-RRT*} used heuristics to improve upon an initial solution generated by RRT. However 

% Learning Approaches
Recent approaches use learning models for non-uniform sampling. \cite{c1}-\cite{c6}. Some of them propose to find low dimensional structure in planning \cite{c1}, \cite{c2}. In particular, generative models like conditional variational autoencoder (CVAE) \cite{c19} have been used to great success. We use the CVAE used in LEGO \cite{c1} (called \legocvae hereon) as our underlying learning model in \gcs and provide a gist of the same below. Note that, although our work uses the model used in LEGO, it can be extended to work with any other learning framework that predicts samples in bottleneck regions \cite{c2}-\cite{c5}.

\subsection{LEGO: Leveraging Experience using Graph Oracles}

Leveraging Experience using Graph Oracles is a framework for predicting efficient roadmaps for sampling based motion planning. During training time, LEGO processes a dense graph to identify a sparse subset of key vertices. These vertices are a diverse set of nodes lying on bottleneck regions through which a near optimal path may pass. A CVAE \cite{c16} conditioned on the occupancy grid of the workspace, start and goal positions is then trained on these key samples. During test time, given the occupancy grid, start and goal positions, the CVAE generates these key samples which are used in conjunction with a uniform sparse graph to generate a roadmap.

\subsection{CVAE: Conditional Variational Autoencoder}

The core component of the LEGO framework is a conditional variational autoencoder (CVAE). It is an extension of traditional variational autoencoder and has been increasingly used to learn low dimensional structure in planning. The addition of conditional parameters helps embed the features of a planning problem as conditions and learn the corresponding representations.

Let $\stateSpace$ be the state space and $\state \in \stateSpace$, $\latentVar \in$ $\mathbb{R} ^ \latentDim$ be the latent random variable and $\condVar \in$ $\mathbb{R}^\featDim$ be the conditioning variable. The framework comprises of two \emph{deterministic mappings} - an \emph{encoder} and a \emph{decoder}. An encoder maps $(\stateVar, \condVar)$ to a mean and variance value of a Gaussian $\encoder(\latentVar | \stateVar, \condVar)$ in latent space, such that it is ``close'' to a standard Gaussian $\Normal(0, I)$. The decoder maps this Gaussian and $y$ to a distribution in the output space $\decoder(\stateVar | \latentVar, \condVar)$. This is achieved by maximizing the following function $\loss \left( \stateVar, \condVar; \decoderParam, \encoderParam \right)$:
\begin{equation}
\label{eq:cvae_loss}
	  -D_{KL}\left( \encoder(\latentVar | \stateVar, \condVar) \, || \, \Normal(0, I) \right) + \frac{1}{\latentDim} \sum_{l=1}^{\latentDim} \log \decoder(\stateVar | \condVar, \latentVar^{(l)})
\end{equation}
At test time, we use only the decoder to map samples from an isotropic Gaussian in the latent space to samples in the output space. The CVAE is trained by passing in a dataset $\dataSet = \{ \stateSetVar_i, \feat_i \}_{i=1}^{D}$. $\feat_i$ is the conditional parameter vector extracted from the planning problem. In our case it's (start, goal, occupancy grid). $\stateSetVar_i$ is the desired set of nodes extracted from the dense graph $\denseGraph$ that we want our learner to predict. Hence we train the model by maximizing the following objective. 
\begin{equation}
	\risk(\dataSet; \decoderParam, \encoderParam) = \frac{1}{\abs{\dataSet}} \sum_{i=1}^{\abs{\dataSet}} \sum_{j=1}^{\abs{\stateSetVar_i}} \loss \left( \stateVar_j, \condVar_i; \decoderParam, \encoderParam \right)
\end{equation}

\section{Problem Formulation}

% This section describes the notations used in this paper and formally defines the motion planning problem addressed by the proposed method.

Let $\stateSpace$ denote a $\Dim-$dimensional configuration space. Let $\obsStateSpace$ $\subseteq$ $\stateSpace$ be the portion in collision and $\freeStateSpace = \stateSpace \setminus \obsStateSpace$ denote the free space. Let a path $\Path : [0, 1] \rightarrow \stateSpace$ be a continuous mapping from index to configurations. A path $\Path$ is said to be collision free if $\Path(\ind) \in \freeStateSpace$ for all $\ind \in [0, 1]$. 

% We define a \emph{motion planning problem} $\pp = \set{\start, \goal, \freeStateSpace}$ as a tuple of start configuration $\start \in \freeStateSpace$, goal configuration $\goal \in \freeStateSpace$ and free space $\freeStateSpace$. 

% Given a problem, a path $\Path$ is said to be \emph{feasible} if it is collision free, $\Path(0)=\start$ and $\Path(1)=\goal$. 
% We wish to solve the motion planning problem of finding a feasible $\Path$. 

\textbf{Problem:} Given a \emph{motion planning problem} $\pp = \set{\start, \goal, \freeStateSpace}$ where $\start \in \freeStateSpace$ is the start configuration and $\goal \in \freeStateSpace$ is the goal configuration, find a \emph{feasible} $\Path$ i.e. $\Path$ that is collision free, $\Path(0)=\start$ and $\Path(1)=\goal$.

Given a database of prior worlds, the overall goal is to use a conjunction of a learned policy and local sampling based planners to generate a roadmap $\graph$ which is used by a graph search algorithm $\alg$ to efficiently compute a feasible path. 
$\alg$, given a graph $\graph$, finds and returns a feasible path $\Path$. If no feasible path exists in the graph, $\alg$ returns $\emptyset$. An ideal roadmap should be sparse enough for $\alg$ to be efficient. In addition, for problems with narrow passages (bottleneck regions), this roadmap must have a set of critical samples in the bottleneck regions to ensure existence of a feasible path.

We assume the graphs to be undirected for simplicity. However, it can easily be extended to directed graphs. The following are the additional notations used in this paper:

\begin{itemize}
  \item $\sparseGraph$ : A sparse graph embedded in the state space of the robot. It is additionally composed with the set of samples generated from the learner and local planners while constructing $\graph$ by $\lcsrrt$ to ensure a minimal coverage. 
  \item $\cs$ : Set of critical sources. 
%   \item $\iscollisionfree$ : A function that takes input a vertex or an edge and returns true if it completely $ \in \freeStateSpace$. Else, it returns false.  
\end{itemize}

\section{Approach}

% It is almost impossible to feed all the fine information to the model regarding the structure of the environment and therefore the complete reliance on learner to exploit local structure can reduce it's generality and can sometimes even make it computationally intractable. 
We propose to identify the location of bottleneck regions using \gcs and then use \lcsrrt or \csrrt to navigate through these bottlenecks using local sampling based planners instead of learning their local structures.

\begin{algorithm}[t]
  \caption{ \gcs }\label{}
%   \algorithmStyle
  \SetKwInOut{Input}{Input}
  \SetKwInOut{Output}{Output}

  \Input{ Planning problem $\pp$, \\ \ Sparse graph $\sparseGraph$}
  \Output{ Set of Critical Sources $\cs$}

  $ \ccs \gets \legoglobal$\;
  $\cs \gets \emptyset$\; 
  \For{sample $\in \ccs$ }
  {
    isNear $\gets$ false\;
    \For{ source $ \in \cs$ }
    {
      \uIf{ dist(source,sample) < $source\_sep$  }
      {
         isNear $\gets$ true \;
      }
    }
    \uIf{ isNear = false }
    {
        free\_count $\gets$ 0 \;
        total\_count $\gets$ 0 \;
        \For{ v $\in \sparseGraph$}
        {
            \uIf{dist(sample,v) < $r\_critical$}
            {
                \uIf{\iscollisionfree(edge(sample,v))}
                {
                    free\_count $\gets$ free\_count $+$ 1 \;
                }
                total\_count $\gets$ total\_count $+$ 1 \;
            }    
        }        
        \uIf{free\_count/total\_count < $threshold$ }
        {
            $ \cs \gets \cs$ $\cup $ sample\;
        }
    }
  }
  \Return{$\cs$}\;
\end{algorithm}
% \vspace{-5mm}
% \vspace{-1cm}
\subsection{Critical Source Generation}

$\gcs$ identifies the locations of bottleneck regions by generating \cs, a set of critical samples, one corresponding to each bottleneck region. It uses $\legocvae$ as the underlying learner. 
% \rkjnote{ASK AVK}
If an occupancy grid of a certain size is required to learn the locations of the bottleneck regions along with their local structures, we argue that we can preprocess this occupancy grid with a kernel to create a grid of lower dimensions that encodes only the information necessary to learn the locations of the bottleneck regions. 
The choice of kernel depends on the features of the environment. For example, a dilation kernel can be used in environments which contain extended narrow passages. 
As we do not expect $\legocvae$ to learn the local structures, there is no loss of relevant information.

% For instance in a 2d environment, for an initial occupancy grid of $N$ by $N$ we can preprocess this occupancy grid with a kernel of size $M$ by $M$ (M > 1) with stride $K$ to extract the necessary information into a grid of size $(floor((N-M/K)+1)$ by $(floor((N-M)/K)+1)$. 

This preprocessed occupancy grid is then used for training and testing $\legocvae$. 
% The decrease in size of the occupancy grid makes it possible for $\legocvae$ to quickly converge to generalizing well over planning problems.
% that have similar structure for the location of bottleneck regions but different local structures within these regions. 
%  \rvnote{change it, robustness thing}.
We call the $\legocvae$ trained with this preprocessed occupancy grid $\legoglobal$ as it only learns global information of the planning problem.
% $\legoglobal$ quickly converge to generalizing well over planning problems due to the decrease in size of the occupancy grid.

We thus use $\legoglobal$ for generating a set of candidate critical source samples. A sample from this set $ \in \cs$ if
(a) it is at least a distance $source\_sep$ away from all the critical sources generated up till now (Lines 4-8, Algorithm 1) and (b) if we connect the sample to the vertices of $\sparseGraph$ that are within distance $r\_critical$ by edges, the  percentage of edges not in collision with the obstacles must be smaller than a certain $threshold$ (Lines 9-17), Algorithm 1).

% \begin{itemize}
%     \item The sample must be at least a distance $source\_sep$ from all the critical sources generated up till now (Lines 4-8, Algorithm 1)
%     \item If we connect the sample to the vertices of $\sparseGraph$ that are within $r$ distance by edges, the  percentage of edges not in collision with the obstacles must be smaller than a certain $threshold$ (Lines 9-17), Algorithm 1).
% \end{itemize}
% Here, $source\_sep$, $r$ and $threshold$ are hyperparameters. 

\begin{algorithm}[t]
  \caption{ \lcsrrt }
%   \algorithmStyle
  \SetKwInOut{Input}{Input}
  \SetKwInOut{Output}{Output}

  \Input{ Planning problem $\pp$, \\ \ Sparse graph $\sparseGraph$}
  \Output{ Path $\Path$}
    $ \cs \gets \gcs(\pp,\sparseGraph) $\;
    $ \cs \gets \cs $ $\cup$ \{start,goal\} \;
    $ \graph \gets \sparseGraph$ $\cup$ $\cs$ \; 
    $ \cstree = \{ \cstree_{1} = \{\cs_{1}\}, \cdots, \cstree_{N} = \{\cs_{|\cs|}\}\}$ \;
    $ \localgraph = \{ \localgraph_{1} = \{\cs_{1}\}, \cdots, \localgraph_{N} = \{\cs_{|\cs|}\}\}$ \;
    $r \gets r\_init$ \;
    \While{True}
    {
      $\localgraph$$\gets$\expandlocalgraphs($\pp,\graph,\localgraph,\cstree,r$)\;
      ($\localgraph,\cstree$)$\gets$\densifylocalgraphs($\localgraph,\cstree,r$) \;
    %   $ \graph \gets \emptyset$ \\
      \For{$i = 1, \cdots, |\cs|$} 
      {
        $ \graph \gets \graph$ $\cup$ $\localgraph_{i}$\;
      }
      $\Path$  = $\alg(\graph,\pp)$\; 
      \uIf{$\Path$ $\neq$ $\emptyset$}
      {
        \Return{$\Path$}\;
      }
      r $\gets \lambda$ r \;
    }
\end{algorithm}
\newcommand{\oldthealgocf}{\arabic{algocf}}
\renewcommand{\thealgocf}{2a}

\begin{algorithm}[t]
  \caption{ \expandlocalgraphs }
%   \algorithmStyle
  \SetKwInOut{Input}{Input}
  \SetKwInOut{Output}{Output}

  \Input{ Planning problem $\pp$, Graph $\graph$, \\ \ Local Graphs Set $\localgraph$ , Trees Set $\cstree$, \\ \ Radius $r$ }
  \Output{ Local Graphs Set $\localgraph$ }
    \For{$i = 1, \cdots, |\localgraph|$} 
    {
      sp $\gets \emptyset$; \Comment{Nodes of $\graph$ within r distance of $\cs_{i}$}\\
      \For{node $ \in \graph$ }
      {
        \uIf{dist($CS_{i}$,node) < $r$}
        {
            sp $\gets$ sp $ \cup$ \{node\}\; 
        }    
      }
      $\localgraph_{i} \gets \localgraph_{i}$ $\cup$ $\subgraph$($\graph$,sp)\; 
      $\setln$ $\gets$ $\localgraph_{i}$ - \connectedcomponent($\localgraph_{i}$,$\cstree_{i}$) \;
      \For{node $\in$ $\setln$ }
      {
        \For{ v $\in$ $\cstree_{i}$}
        {
            \uIf{\iscollisionfree(edge(node,v))}
            {
                $\setln$ $\gets$ $\setln$ - \connectedcomponent($\localgraph_{i}$,node)\;
                $\localgraph_{i} \gets \localgraph_{i}$ $\cup$ \{edge(node,v)\}\;
                % \uIf{$\setln = \emptyset$}
                % {
                %     goto 1\;
                % }
            }  
        }
      }       
    }
  \Return{$\localgraph$}\;
\end{algorithm}
\renewcommand{\thealgocf}{2b}

\begin{algorithm}[t]
  \caption{ \densifylocalgraphs }
%   \algorithmStyle
  \SetKwInOut{Input}{Input}
  \SetKwInOut{Output}{Output}
  \SetKwFor{RepTimes}{repeat}{ until $\setln = \emptyset$ or M iterations}

  \Input{ Local Graphs Set $\localgraph$ , Trees Set $\cstree$, \\ \ Radius $r$}
  \Output{ Local Graphs Set $\localgraph$ , Trees Set $\cstree$}
   \For{$i = 1, \cdots, |\localgraph|$} 
    {
        \uIf{$\localgraph_{i}$ is not connected}
        {
          $\setln$$\gets$$\localgraph_{i}$ -$\connectedcomponent$($\localgraph_{i}$,$\cstree_{i}$)\;
          \RepTimes{}
          {
            \Repeat{Edge (nn,rn') is not in collision }
            {
              rn $\gets$ \randomnode($\cs_{i}$,$r$)\;
              nn $\gets$ \nearestvertex($\cstree_{i}$,rn)\;
              rn'$\gets$ \interpolate(nn, rn, step\_size)\;
            }
            $\cstree_{i} \gets \cstree_{i}$ $\cup$ \{rn',edge(rn',nn)\}\;
            $\localgraph_{i} \gets \localgraph_{i}$ $\cup$ \{rn',edge(rn',nn)\}\;
            \For{n $\in$ $\setln$}
            {
              \uIf{$\iscollisionfree$(edge(rn',n))}
              {
                $\setln$$\gets$ $\setln$ - $\connectedcomponent$($\localgraph_{i}$,n)\;
                $\localgraph_{i} \gets \localgraph_{i}$ $\cup$ \{edge(rn',n)\}\;
              }
            }   
          }
       }
    }    
  \Return{$\localgraph, \cstree$}\;
\end{algorithm}

\subsection{Local Critical Source - RRT}

Local Critical Source - RRT (LCS-RRT) uses \gcs to get \cs and adds the start and goal vertices to it. It builds $\graph$, which contains $\sparseGraph$ and $\cs$. For each $\cs_{i}$, the algorithm maintains two graphs: a tree $\cstree_{i}$  and a local graph $\localgraph_{i}$. $\cstree_{i}$ is RRT rooted at $\cs_{i}$. $\localgraph_{i}$ consists of edges and vertices of $\graph$ within distance $r$ from $\cs_{i}$. The goal of the algorithm is to make $\localgraph_{i}$ completely connected by adding edges between the vertices of $\cstree_{i}$ and the vertices of $\localgraph$ not belonging to the connected component of $\cs_{i}$. 

Initially, for each $\cs_{i}$, both $\cstree_{i}$ and $\localgraph_{i}$ contain only $\cs_{i}$. In an iteration of the outer while loop of LCS-RRT (Line 7, Algorithm 2), $\expandlocalgraphs$ expands each $\localgraph_{i}$ to radius $r$. After expansion, these $\localgraph{i}$ consist of the subgraph of $\graph$ within a radius $r$ of $\cs_{i}$ (Line 6, Algorithm 2a). $\expandlocalgraphs$ and $\densifylocalgraphs$ both maintain $\setln$, a set of vertices of the $\localgraph_{i}$ that does not belong to the connected component of $\cs_{i}$. Wherever possible, the nodes of $\cstree_{i}$, are connected to $\setln$ by collision free edges and $\setln$ is updated (Lines 8-12, Algorithm 2a). $\densifylocalgraphs$ densifies the local graph by growing its $\cstree_{i}$ and adding edges between its $\cstree_{i}$ and the vertices of the set $\setln$ until either the $\localgraph_{i}$ is completely connected or M (hyperparameter) iterations have taken place, whichever is earlier. 
To grow its $\cstree_{i}$, $\densifylocalgraphs$ samples a random node within a radius $r$ of $\cs_{i}$, interpolates it to within a distance of step\_size of the vertex nearest to this node in its $\cstree_{i}$ and, if possible, joins them with an edge in a similar fashion to RRT growth (and therefore the name) (Lines 5-10 Algorithm 2a). It then tries to connect this new node of $\cstree_{i}$ with the vertices of $\setln$ and wherever a connection is possible, $\setln$ is updated (Lines 12-15 Algorithm 2b). After densification, $\localgraph$ is added to $\graph$ and $\alg$ is run on G (Lines 10-12 Algorithm 2). If successful, the feasible $\Path$ is returned. Else, the radius $r$ is increased by a factor $\lambda$ and the process is repeated again and so on. 

\textit{Probabilistic Completeness:} Let us consider the local graph $\localgraph_{s}$ at start node $\cs_{s}$ at infinite time. As time approaches infinity, so does the radius $r$. Thus, the goal node is present in the $\localgraph_{s}$. As we try to connect $\cs_{s}$ to all nodes not present in the connected component of $\cs_{s}$ using $\cstree_{s}$, we are at the least running a RRT from start node to goal node. Thus, due to the probabilistic completeness of RRT, $\lcsrrt$ is probabilistically complete. 

\subsection{Critical Source - RRT}

Critical Source - RRT (CS-RRT) uses \gcs to get $\cs$ and adds them to $\graph$. It then adds start and goal vertices to $\graph$. These nodes subsequently act as the roots of the RRTs we will grow. In each iteration of the outer while loop (Line 3, Algorithm 3), the algorithm iterates through the connected components present in $\graph$ i.e. the RRTs in a round robin manner. In lines 5-10, the algorithm samples a random node, interpolates it to within a distance of $step\_size$ of the vertex nearest to this node in its RRT and loops until it is possible to join them with an edge. This is similar to how a RRT grows (and therefore the name). In lines 11-13 it then tries to connect the new\_node of its tree to the nearest vertices of the other trees that lie within a distance of $step\_size$. If a connection is possible, an edge is inserted into the graph (line 14). Insertion of this edge results in merger of the two trees the nodes belonged to. The algorithm returns a feasible path when the start and goal nodes belong to the same connected component of $\graph$ (Lines 15-17). 

\textit{Probabilistic Completeness:} As $\csrrt$ grows RRTs rooted at start and goal nodes along with the critical sources to connect the start and goal nodes, it runs a RRT-Connect at the least. Thus, due to the probabilistic completeness of RRT-Connect, $\csrrt$ is probabilistically complete. 

\renewcommand{\thealgocf}{3}
\begin{algorithm}[t]
  \caption{ \csrrt }
%   \algorithmStyle
  \SetKwInOut{Input}{Input}
  \SetKwInOut{Output}{Output}

  \Input{ Planning problem $\pp$, \\ \ Sparse graph $\sparseGraph$}
  \Output{ Path $\Path$}
    $ \graph \gets \gcs(\pp,\sparseGraph) $\;
    $ \graph \gets \graph$ $\cup$ \{start,goal\} \;
    \While{True}
    {
      \For{connected component $\graph_{i}$ \(\in\) $\graph$}
      {
        \Repeat{\iscollisionfree(edge(nn,rn')}
        {
          rn $\gets$ \randomnode\;
          nn $\gets$ \nearestvertex($G_{i}$,rn)\;
          rn'$\gets$ \interpolate(nn, rn, $step\_size$)\;
        }
        $ \graph \gets$ $\cup$ \{rn',edge(rn',nn)\}\;
        \For{connected component $\graph_{j} \ne \graph_{i} \in \graph$}
        {
            onn $\gets$ \nearestvertex($\graph_{j}$,rn')\;
            \uIf{distance(onn,rn') < $step\_size$ and \iscollisionfree(edge(onn,rn')) }
            {
                $ \graph \gets \graph$ $\cup$ \{edge(rn',onn)\}\;
                \uIf{Start and Goal belong to same connected component of $\graph$ }
                {
                    $\Path$  = $\alg(\graph,\pp)$\; 
                    \Return{$\Path$}\;
                }
            }       
        }
      }
    }
\end{algorithm}
% \setlength{\textfloatsep}{-2cm}
% \setlength{\floatsep}{-2cm}

% \subsection*{Proof of Probabilistic Completeness}
% \paragraph{$\lcsrrt$}
% Let us consider the local graph $\localgraph_{s}$ at start node $\cs_{s}$ at infinite time. As time approaches infinity, so does the radius for the construction of $\localgraph_{s}$. Thus, the goal node is present in the $\localgraph_{s}$ of $\cs_{s}$. As we try to connect $\cs_{s}$ to all nodes not present in the connected component of $\cs_{s}$ using $\cstree_{s}$, we are at the least running a RRT from start node to goal node. Thus, due to the probabilistic completeness of RRT, $\lcsrrt$ is probabilistically complete. 
% \paragraph{$\csrrt$}
% As $\csrrt$ grows RRTs rooted at start and goal nodes along with the critical sources to connect the start and goal nodes, it runs a RRT-Connect at the least. Thus, due to the probabilistic completeness of RRT-Connect, $\csrrt$ is probabilistically complete. 

\section{Experiments}

In this section, we compare the performance of LCS-RRT and CS-RRT to sampling based algorithms in multiple domains. We evaluate our algorithms against RRT-Connect\cite{c18}, a variation of RRT that incrementally builds two trees rooted at the start and goal nodes. Additionally, we compare our algorithms with LEGO, a state-of-the-art learning based sampling algorithm. 
\begin{figure}[!ht]
    \centering
	\begin{subfigure}[h]{0.32\columnwidth}
		\centering
		  \includegraphics[width=\textwidth, frame]{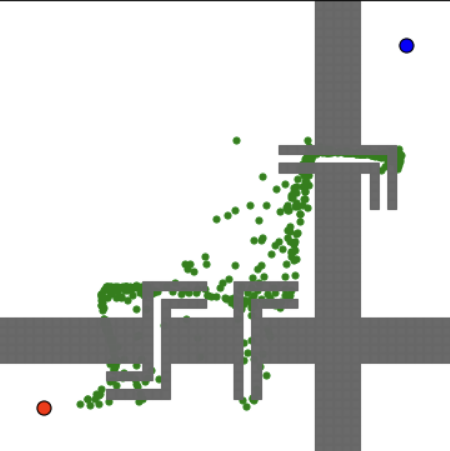}
		  \caption{}
	\end{subfigure}
	\hfill
	\begin{subfigure}[h]{0.32\columnwidth}
		\centering
		  \includegraphics[width=\textwidth, frame]{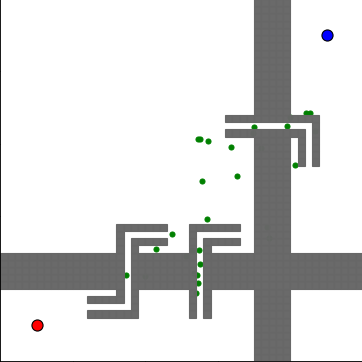}
		  \caption{}
	\end{subfigure}
	\hfill
	\begin{subfigure}[h]{0.32\columnwidth}
		\centering
		  \includegraphics[width=\textwidth, frame]{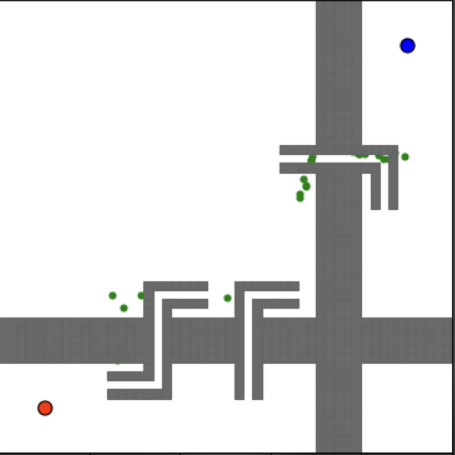}
		  \caption{}
	\end{subfigure}
	\caption{(a) \legocvae fails to cover the local structure of the bottleneck regions even with a large number of samples. (b) \legoglobal is able and (c) \legocvae is unable to place at least one sample per bottleneck region with a small number of samples. Trained with a preprocessed grid, \legoglobal is robust to fine changes in local structures. }
\end{figure}
As LEGO is a graph based approach while others are tree based approaches, we adapted LEGO to an anytime algorithm with incremental densification for testing purposes. 
% Both RRT-Connect and LEGO have been tuned to their best possible parameters.
We have tuned the parameters of RRT-Connect and LEGO to ensure their best performance. 

\paragraph{Evaluation Procedure}
For a given planning problem, we run each algorithm with a fixed timeout. For a given problem domain, each of the learning models used by these algorithms are trained on similar number of planning problems. We evaluate the performance of these algorithms on the metric of time taken to find a feasible path. 

\paragraph{Problem Domains}
We evaluate our algorithms on $\mathbb{R}^2$ and $\mathbb{R}^7$ problem domains. The $\mathbb{R}^2$ problems have random rectilinear walls with extruded narrow passages that have varying local structures (Fig 3).
%The width of the narrow passages may be small (2D Hard), medium (2D Medium) or large (2D Hard).
The $\mathbb{R}^7$ problem is a robotic-arm manipulation problem in a cluttered environment.

\paragraph{Experiment Details}
We compare the algorithms $\lcsrrt$, $\csrrt$, LEGO and RRT-Connect on a testset of 100 planning problems. 
% We run these algorithms 10 times on each of these planning problems and take their mean to reduce randomness \rvnote{should we keep this?}. 
For the $\mathbb{R}^2$ problems, each learning model is trained on 4000 planning problems for around 30 minutes. The planning timeout for the algorithms is 5 seconds. The size of occupancy grid used by $\legocvae$ is 50X50. The kernel used by $\legoglobal$ for preproccessing is of size 5X5 (with stride value 5) resulting in an updated occupancy grid of size 10X10 (Fig 5). For the $\mathbb{R}^7$ problems, each learning model is trained on 4000 training problems for 2 hours and the planning timeout for the algorithms is 12 seconds.
% The hyperparameters of the algorithms and the architecture of the models used have been described in the APPENDIX. 
The code is open sourced and can be found at \url{https://github.com/RKJenamani/CS-RRT}. 

% \subsection*{Observations}
% \newtheorem{observation}[thm]{O}
% \begin{observation}
% \textit{Observation} \\
\textbf{Observation 1.} \textit{\lcsrrt and \csrrt outperform the sampling based baselines LEGO and RRT-Connect.} \\
% \end{observation}
% \textbf{O 1: } 
Fig 4 shows the performance of $\csrrt$, $\lcsrrt$, LEGO and RRT-Connect on $\mathbb{R}^2$ and $\mathbb{R}^7$ problem domains. $\lcsrrt$ performs better than $\csrrt$ in $\mathbb{R}^2$. 
However in $\mathbb{R}^7$, where the size of $\sparseGraph$ becomes large to ensure coverage of the high dimensional space and collision checking is computationally expensive, $\csrrt$ performs better. \\
% \begin{observation}
\textbf{Observation 2.} \textit{\legocvae is unable to cover the local structure of the bottleneck regions even with a large number of samples. Also, with a few number of samples, \legoglobal is able to place a sample at each bottleneck region while \legocvae is not (Fig 2)}.

\begin{figure*}[!ht]
	\centering
% 	\begin{subfigure}[h]{0.45\columnwidth}
% 		\centering
% 		  \includegraphics[width=\textwidth, frame]{figs/env.png}
% 		  \caption{}		
% 	\end{subfigure}
% 	\hfill
% 	\begin{subfigure}[h]{0.45\columnwidth}
% 		\centering
% 		  \includegraphics[width=\textwidth, frame]{figs/env_for_LEGO.png}
% 		  \caption{}		
% 	\end{subfigure}
% 	\hfill
% 	\begin{subfigure}[h]{0.45\columnwidth}
% 		\centering
% 		  \includegraphics[width=\textwidth, frame]{figs/samples_by_LEGO.png}
% 		  \caption{}		
% 	\end{subfigure}
% 	\hfill
	\begin{subfigure}[h]{0.45\columnwidth}
		\centering
		  \includegraphics[width=\textwidth]{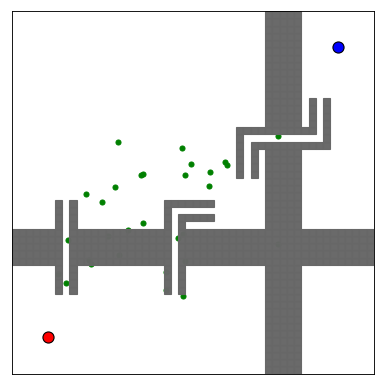}
		  \caption{}		
	\end{subfigure}
		\hfill
	\begin{subfigure}[h]{0.45\columnwidth}
		\centering
		  \includegraphics[width=\textwidth]{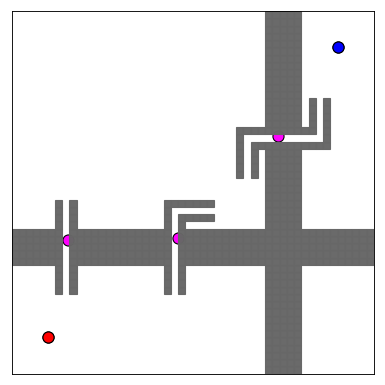}
		  \caption{}		
	\end{subfigure}
	\hfill
	\begin{subfigure}[h]{0.45\columnwidth}
		\centering
		  \includegraphics[width=\textwidth]{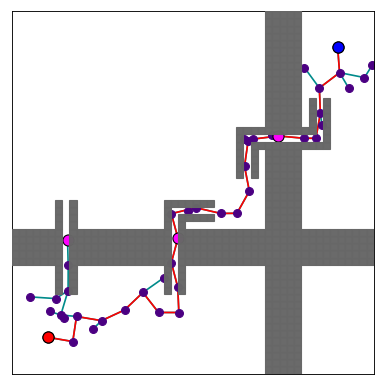}
		  \caption{}		
	\end{subfigure}
	\hfill
	\begin{subfigure}[h]{0.45\columnwidth}
		\centering
		  \includegraphics[width=\textwidth]{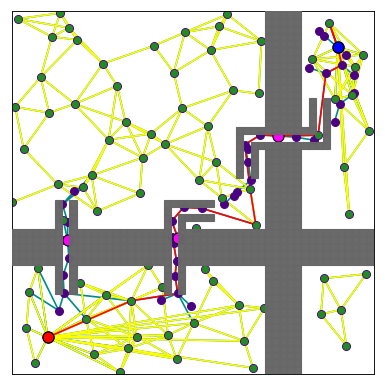}
		  \caption{}		
	\end{subfigure}
	\caption{(a) Samples predicted by \legoglobal(green) on the $\mathbb{R}^2$ environment. (b) Critical Sources (Pink) selected by \gcs. (c) Path (Red) found by \csrrt (d) Path (Red) found by \lcsrrt }
	\label{fig:envs}
\end{figure*}
\begin{figure}[!ht]
    \centering
	\begin{subfigure}[h]{0.48\columnwidth}
		\centering
		  \includegraphics[width=\textwidth]{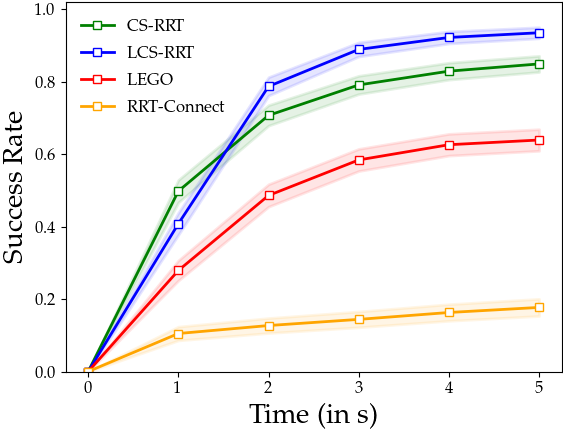}
		  \caption{}
	\end{subfigure}
	\hfill
	\begin{subfigure}[h]{0.48\columnwidth}
		\centering
		  \includegraphics[width=\textwidth]{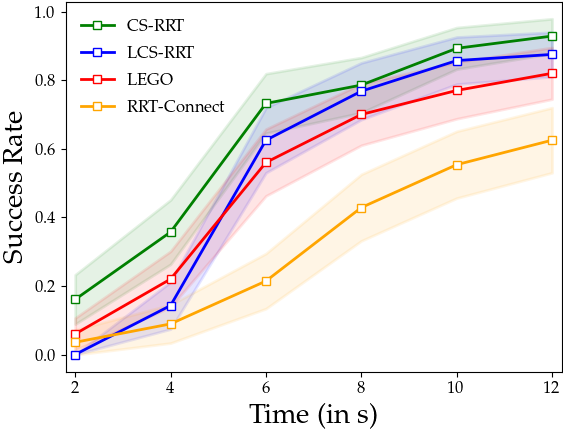}
		  \caption{}
	\end{subfigure}
	\caption{Comparison of CS-RRT, LCS-RRT, RRT-Connect and LEGO for (a) $\mathbb{R}^2$ and (b) $\mathbb{R}^7$ problem domains}
\end{figure}
\begin{figure}[!ht]
    \centering
	\begin{subfigure}[h]{0.48\columnwidth}
		\centering
		  \includegraphics[width=\textwidth]{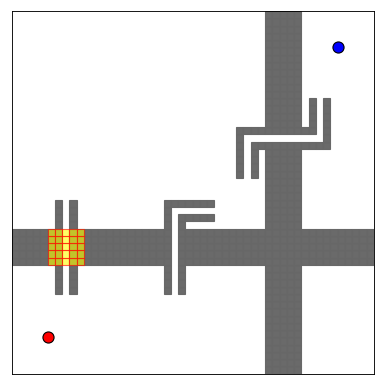}
		  \caption{}
	\end{subfigure}
	\hfill
	\begin{subfigure}[h]{0.48\columnwidth}
		\centering
		  \includegraphics[width=\textwidth]{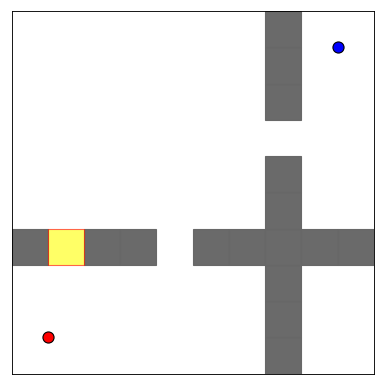}
		  \caption{}
	\end{subfigure}
	\caption{Convolving with a kernel of size 5X5 on (a) an occupancy grid of size 50 by 50, we are able to extract (b) a 10 by 10 occupancy grid with global features. This approach is similar to the method of dilation used in image processing. }
\end{figure}

% \end{observation}
% \textbf{O 2: }

% \textbf{O 3: }

% \textbf{O 3: }\lego outperforms 

% \begin{observation}
 
% \end{observation}

\section{Conclusion}

We show the feasibility of using local sampling algorithms aided by a learning model to rapidly find a feasible path in complex environments containing extended bottleneck regions. These algorithms share the responsibility of generating key samples between the learner and the local sampler. This lets the learning model converge faster to generalizing well over planning problems that have similar global structure for the location of bottleneck regions but different local structures within these regions. As we require the learner to only identify the location of bottleneck regions, we introduce the idea of using a kernel to preprocess the occupancy grids for better learning. In future works, we would like to analyse the relationship between a workspace and the kernel suitable to it. We intend to explore the integration of other sampling based methods to make the approach asymptotically optimal. We would also like to test our algorithms on environments where extended bottleneck regions arise due to differential constraints. 
% \section*{ACKNOWLEDGMENT}

% We would like to thank Manjunath Bhat and Aditya Vamsikrishna Mandalika for th


\begin{thebibliography}{99}

\bibitem{c1} R. Kumar, A. Mandalika, S. Choudhury, and S. S. Srinivasa, “LEGO: Leveraging experience in roadmap generation for sampling-based planning”, in IROS 2019.
\bibitem{c2} B. Ichter, J. Harrison, and M. Pavone, “Learning sampling distributions for robot motion planning”, in ICRA 2018.
\bibitem{c3} B. Ichter, E. Schmerling, T.-W. E. Lee, and A. Faust, "Learned Critical Probabilistic Roadmaps for Robotic Motion Planning", arXiv preprint arXiv:1910.03701, 2019. 
\bibitem{c4} C. Chamzas, A. Shrivastava, Lydia E. Kavraki, "Using Local Experiences for Global Motion Planning", in ICRA 2019.
\bibitem{c5} Daniel Molina, Kislay Kumar, Siddharth Srivastava, “Learn and Link: Learning Critical Regions for Efficient Planning,” in ICRA 2020.
\bibitem{c6} S.R. Koukuntla, M. Bhat, S. Aggarwal, R. K. Jenamani, and J. Mukhopadhyay, "Deep Learning rooted Potential piloted RRT* for expeditious Path Planning", in CACRE 2019.
\bibitem{c7} D. Hsu, J.-C. Latombe, and R. Motwani, "Path planning in expansive configuration spaces", in International Journal Computational Geometry and Applications, 4:495–512, 1999.
\bibitem{c8} L. Janson, B. Ichter, and M. Pavone, "Deterministic sampling-based motion planning: Optimality, complexity, and performance", arXiv preprint arXiv:1505.00023, 2015.
\bibitem{c9} Christopher Holleman and Lydia E. Kavraki, "A framework for using the workspace medial axis in PRM planners", in ICRA 2000.
\bibitem{c10} D. Hsu, G. Sánchez-Ante, and Z. Sun, "Hybrid prm sampling with a cost-sensitive adaptive strategy", in ICRA 2005.
\bibitem{c11} Brendan Burns and Oliver Brock, "Sampling-based motion planning using predictive models", in ICRA 2005.
\bibitem{c12} D. Hsu, J. Latombe, and H. Kurniawati, "On the probabilistic foundations of probabilistic roadmap planning",  in IJRR 2006.
\bibitem{c13} H. Kurniawati and D. Hsu, "Workspace-based connectivity oracle: An adaptive sampling strategy for prm planning" in Algorithmic Foundation of Robotics VII, pages 35–51. Springer, 2008.
\bibitem{c14} Valérie Boor, Mark H. Overmars, and A. Frank Van Der Stappen, "The gaussian sampling strategy for probabilistic roadmap planners" in ICRA 1999.
\bibitem{c15} D. Hsu, T. Jiang, J. Reif, and Z. Sun, "The bridge test for sampling narrow passages with probabilistic roadmap planners", in ICRA 2003.
\bibitem{c16} Diederik P. Kingma and Max Welling, "Auto-encoding variational bayes", arXiv preprint arXiv:1312.6114, 2013
\bibitem{c17} S. M. LaValle, "Rapidly-exploring random trees: A new tool for path planning". TR 98-11, Computer Science Dept., Iowa State Univ. <http://janowiec.cs.iastate.edu/papers/rrt.ps>, Oct. 1998.
\bibitem{c18} J. Kuffner and S. M. LaValle, "RRT-connect: An efficient approach to single-query path planning”, in ICRA 2000.
\bibitem{c19} Carl Doersch, "Tutorial on variational autoencoders", arXiv preprint arXiv:1606.05908, 2016.
\end{thebibliography}
\end{document}